\documentclass[11pt,a4paper]{article}
\usepackage[hyperref]{emnlp-ijcnlp-2019}
\usepackage{times}
\usepackage{latexsym}

\usepackage{url}

\usepackage{amsmath}
\usepackage{amssymb}

\usepackage{graphicx}
\usepackage{tikz}
\usepackage{tikz-dependency}
\usepackage{CJKutf8}

\aclfinalcopy 

\title{Syntax-Enhanced Self-Attention-Based Semantic Role Labeling}

\author{Yue Zhang,  Rui Wang, Luo Si \\
Alibaba Group, China  \\
\{\tt shiyu.zy, masi.wr, luo.si\}@alibaba-inc.com}

\date{}

\begin{document}
\maketitle
\begin{abstract}
As a fundamental NLP task, semantic role labeling (SRL) aims to discover the semantic roles for each predicate within one sentence. 
This paper investigates how to incorporate syntactic knowledge into the SRL task effectively. 
We present different approaches of encoding the syntactic information derived from dependency trees of different quality and representations; we propose a syntax-enhanced self-attention model and compare it with other two strong baseline methods; and we conduct experiments with newly published deep contextualized word representations as well. The experiment results demonstrate that with proper incorporation of the high quality syntactic information, our model achieves a new state-of-the-art performance for the Chinese SRL task on the CoNLL-2009 dataset.
\end{abstract}

\section{Introduction}
\begin{figure}[t]
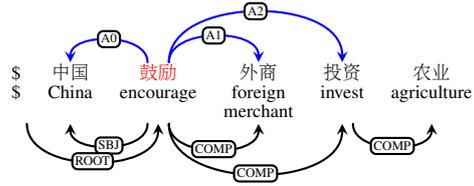

\centering
\scriptsize
\begin{dependency}[arc edge, arc angle=90]
\begin{deptext}[column sep=1.0em]
\$ \& 中国 \&  {\color{red}  鼓励} \& 外商 \& 投资 \& 农业  \\
\$ \& China \& encourage \& foreign  \& invest \& agriculture  \\
 \&  \&  \& merchant \&  \&  \&   \\
\end{deptext}

\depedge[edge below, edge style={black, thick}]{3}{2}{SBJ}
\depedge[edge below, edge style={black, thick}]{1}{3}{ROOT}
\depedge[edge below, edge style={black, thick}]{3}{4}{COMP}
\depedge[edge below, edge style={black, thick}]{3}{5}{COMP}
\depedge[edge below, edge style={black, thick}]{5}{6}{COMP}

\depedge[edge above, edge style={blue, thick}]{3}{2}{A0}
\depedge[edge above, edge style={blue, thick}]{3}{4}{A1}
\depedge[edge above, edge style={blue, thick}]{3}{5}{A2}
\end{dependency}
\caption{An example of one sentence with its syntactic dependency tree and semantic roles. Arcs above the sentence are semantic role annotations for the predicate ``鼓励 (encourage)" and below the sentence are syntactic dependency annotations of the whole sentence. 
The meaning of this sentence is ``China encourages foreign merchants to invest in agriculture".}
\label{graph:example}
\end{figure}

The task of semantic role labeling (SRL) is to recognize arguments for a given predicate in one sentence and assign labels to them, including “who” did “what” to “whom”, “when”, “where”, etc.
Figure \ref{graph:example} is an example sentence with both semantic roles and syntactic dependencies. Since the nature of semantic roles is more abstract than the syntactic dependencies, SRL has a wide range of applications in different areas, e.g., text classification \cite{sinoara2016}, text summarization \cite{genest2011,khan2015}, recognizing textual entailment \cite{burchardt2007,stern2014}, information extraction \cite{surdeanu2003}, question answering \cite{shen2007,yih2016}, and so on. 

\begin{CJK}{UTF8}{gbsn}
\end{CJK}


Traditionally, syntax is the bridge to reach semantics. However, along with the popularity of the end-to-end models in the NLP community, various recent studies have been discussing the necessity of syntax in the context of SRL. For instance, \citet{he2017} have observed that only good syntax helps with the SRL performance. \citet{xia2018} have explored what kind of syntactic information or structure is better suited for the SRL model. \citet{cai2018} have compared syntax-agnostic and syntax-aware approaches and claim that the syntax-agnostic model surpasses the syntax-aware ones.


In this paper, we focus on analyzing the relationship between the syntactic dependency information and the SRL performance. In particular, we investigate the following four aspects: 1) \emph{Quality} of the syntactic information: whether the performance of the syntactic parser output affects the SRL performance; 2) \emph{Representation} of the syntactic information: how to represent the syntactic dependencies to better preserve the original structural information; 3) \emph{Incorporation} of the syntactic information: at which layer of the SRL model and how to incorporate the syntactic information; and 4) the \emph{Relationship} with other external resources: when we append other external resources into the SRL model, whether their contributions are orthogonal to the syntactic dependencies.

For the main architecture of the SRL model, many neural-network-based models use BiLSTM as the encoder (e.g., \citet{cai2018,lstmsrlzhao2018,heluheng2018}), while recently self-attention-based encoder becomes popular due to both the effectiveness and the efficiency \cite{vaswani2017,tan2017,lisa2018}. By its nature, the self-attention-based model directly captures the relation between words in the sentence, which is convenient to incorporate syntactic dependency information. \citet{lisa2018} replace one attention head with pre-trained syntactic dependency information, which can be viewed as a \emph{hard} way to inject syntax into the neural model. Enlightened by the machine translation model proposed by \citet{ra2018}, we introduce the \emph{Relation-Aware} method to incorporate syntactic dependencies, which is a \emph{softer} way to encode richer structural information.

Various experiments for the Chinese SRL on the CoNLL-2009 dataset are conducted to evaluate our hypotheses. From the empirical results, we observe that: 1) The quality of the syntactic information is essential when we incorporate structural information into the SRL model; 2) Deeper integration of the syntactic information achieves better results than the simple concatenation to the inputs; 3) External pre-trained contextualized word representations help to boost the SRL performance further, which is not entirely overlapping with the syntactic information.

In summary, the contributions of our work are:
\begin{itemize}
    \item We present detailed experiments on different aspects of incorporating syntactic information into the SRL model, in what quality, in which representation and how to integrate.
    \item We introduce the relation-aware approach to employ syntactic dependencies into the self-attention-based SRL model.
    \item We compare our approach with previous studies, and achieve state-of-the-art results with and without external resources, i.e., in the so-called \emph{closed} and \emph{open} settings.
\end{itemize}

\section{Related work}
Traditional semantic role labeling task \cite{gildea2002} presumes that the syntactic structure of the sentence is given, either being a constituent tree or a dependency tree, like in the CoNLL shared tasks \cite{carreras2005,surdeanu2008,hajic2009}. Recent neural-network-based approaches can be roughly categorized into two classes: 1) making use of the syntactic information \cite{fitzgerald2015,roth2016,qian2017,marcheggiani2017b}, and 2) pure end-to-end learning from tokens to semantic labels, e.g., \citet{zhou2015,marcheggiani2017a}.

\citet{roth2016} utilize an LSTM model to obtain embeddings from the syntactic dependency paths; while \citet{marcheggiani2017b} construct Graph Convolutional Networks to encode the dependency structure. Although \citet{he2017}'s approach is a pure end-to-end learning, they have included an analysis of adding syntactic dependency information into English SRL in the discussion section. \citet{cai2018} have compared syntax-agnostic and syntax-aware approaches and \citet{xia2018} have compared different ways to represent and encode the syntactic knowledge.

In another line of research, \citet{tan2017} utilize the Transformer network for the encoder instead of the BiLSTM. \citet{lisa2018} present a novel and effective multi-head self-attention model to incorporate syntax, which is called LISA (Linguistically-Informed Self-Attention). We follow their approach of replacing one attention head with the dependency head information, but use a softer way to capture the pairwise relationship between input elements \cite{ra2018}.




For the datasets and annotations of the SRL task, most of the previous research focuses on 1) PropBank \cite{palmer2005} and NomBank \cite{meyers2004} annotations, i.e., the CoNLL 2005 \cite{carreras2005} and CoNLL 2009 \cite{hajic2009} shared tasks; 2) OntoNotes annotations \cite{weischedel2011}, i.e., the CoNLL 2005 and CoNLL 2012 datasets and more; 3) and FrameNet \cite{baker1998} annotations. For the non-English languages, not all of them are widely available. Apart from these, in the broad range of semantic processing, other formalisms non-exhaustively include abstract meaning representation \cite{banarescu2013}, universal decompositional semantics \cite{white2016}, and semantic dependency parsing \cite{oepen2015}. \citet{abend2017} give a better overview of various semantic representations. In this paper, we primarily work on the Chinese and English datasets from the CoNLL-2009 shared task and focus on the effectiveness of incorporating syntax into the Chinese SRL task.

\section{Approaches}

\begin{figure}
\hspace*{-2.5em}
  \centering
  \includegraphics[width=.47\textwidth]{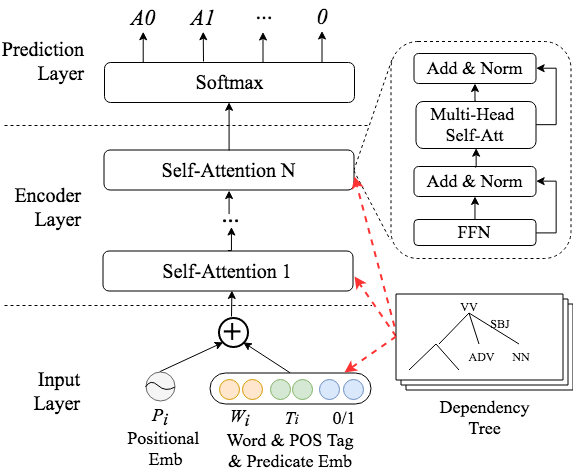} 
  \caption{Architecture of our syntax-enhanced self-attention-based SRL model. Red dotted arrows indicate different locations where we incorporate linguistic knowledge in different forms. The dotted box on the upper right is the detailed composition of the self-attention block.}  
  \label{modelimg}
\end{figure}

In this section, we first introduce the basic architecture of our self-attention-based SRL model, and then present two different ways to encode the syntactic dependency information. Afterwards, we compare three approaches to incorporate the syntax into the base model, concatenation to the input embedding, LISA, and our proposed relation-aware method.

\subsection{The Basic Architecture}
Our basic model is a multi-head self-attention-based model, which is  effective in SRL task as previous work proves \cite{tan2018deep}.
The model consists of three layers: the input layer, the encoder layer and the prediction layer as shown in Figure \ref{modelimg}.

\subsubsection{Input Layer}
The input layer contains three types of embeddings: token embedding, predicate embedding, and positional embedding.

\textbf{Token Embedding} includes word embedding, part-of-speech (POS) tag embedding.    

\textbf{Predicate Embedding} has been proposed by \citet{he2017}, and its binary embedding is used to indicate the predicates indices in each sentence. 

\textbf{Positional Embedding} encodes the order of the input word sequence. We follow \citet{vaswani2017} to use time positional embedding, which is formulated as follows:
\begin{equation}
\begin{split}
    PE(t, 2i) &= sin(t / 10000^{2i/d}) \\
    PE(t, 2i+1) &= cos(t / 10000^{2i/d})
 \end{split}
\end{equation}
where $t$ is the position,  $i$ means the dimension, and $d$ is the dimension of the model input embedding.

\subsubsection{Encoder Layer}
The self-attention block is almost the same as Transformer encoder proposed by \citet{vaswani2017}. Specifically the Transformer encoder contains a feed-forward network (\textbf{FFN}) and a multi-head attention network. 
The former is followed by the latter. 
In this work, we exchange their order, so that the multi-head attention module is moved behind the FFN module\footnote{Changing the order delivers better empirical results \cite{tan2018deep} and our experiments show the same trend. The details of the experiments are not listed in this paper.} as Figure \ref{modelimg} shows. 

\textbf{FFN} The FFN module consists of two affine layers with a ReLU activation in the middle. Formally, we have the following equation:
\begin{equation}
FFN(x) = max(0, xW_1 + b_1)W_2 + b_2
\end{equation}

\textbf{Multi-Head Attention} The basic attention mechanism used in the multi-head attention function is called ``Scaled Dot-Product Attention'', which is formulated as follows:
\begin{equation}
Attention(Q, K, V) = softmax(\frac{QK^T}{\sqrt{d_k}})V
\end{equation}
where $Q$ is queries, $K$ is keys, and $V$ is values. 

In the multi-head attention setting, it first maps the input matrix $X$ into queries, keys and values matrices by using $h$ different learned linear projections. Taking queries $Q$ as an example:
\begin{equation}
Linear_i^{Q}(X) = XW_{i}^{Q} + b_{i}^{Q}
\end{equation}
where $0 \leq i < h$. 
Keys and values use similar projections.

On each of these projections, we perform the scaled dot-product attention in parallel. These parallel output values are concatenated and once again projected into the final values. Equation \ref{eq:multi-head} depicts the above operations.
\begin{equation} \label{eq:multi-head}
\small
MultiHead(X) = Concat(head_1, \ldots , head_h)W^O 
\end{equation}
where 
\begin{equation}
\begin{aligned}
\label{ra}
\scriptsize
head_i = Att&ention(Linear_i^{Q}(X),  \\
 & Linear_i^{K}(X)  , Linear_i^{V}(X))
\end{aligned}
\end{equation}
More details about multi-head attention can be found in \citet{vaswani2017}.

\textbf{Add \& Norm} We employ a residual connection to each module, followed by a layer normalization \cite{Ba-2016-layernorm} operation. 
The output of each module is formulated as 
\begin{equation}
x = LayerNorm(x + f(x))
\end{equation}
where $f(x)$ is implemented by each above module.

\subsection{Representation of the Syntactic Dependencies}

\subsubsection{Dependency Head \& Relation}

The most intuitive way to represent syntactic information is to use individual dependency relations directly, like dependency head and dependency relation label, denoted as \textsc{Dep} and \textsc{Rel} for short.

Except for \textsc{LISA}, where \textsc{Dep} is a one-hot matrix of dependency head word index described in \ref{lisa_description}, in other cases, we use the corresponding head word. \textsc{Rel} is the dependency relation between the word and its syntactic head. 
We take both \textsc{Dep} and \textsc{Rel} as common strings and map them into dense vectors in the similar way of word embedding.

\subsubsection{Dependency Path \& Relation Path }
In order to preserve the structural information of dependency trees as much as possible, we take the syntactic path between candidate arguments and predicates in dependency trees as linguistic knowledge. 
Referring to \citet{xia2018}, we use the Tree-based Position Feature (TPF) as Dependency Path (\textsc{DepPath}) and use the Shortest Dependency Path (SDP) as Relation Path (\textsc{RelPath}).

To generate \textsc{DepPath \& RelPath} between candidate argument and predicate, we firstly find their lowest common ancestor. 
Then we get two sub-paths, one is from the ancestor to the predicate and the other is from the ancestor to the argument.
For \textsc{DepPath}, we compute distance from ancestor to predicate and argument respectively and then concatenate two distances with the separator `,'. 
For \textsc{RelPath}, we concatenate the labels appearing in each sub-path with the separator ``\_" respectively to get two label paths, and then concatenate the two label paths with the separator `,'. 

\begin{figure}[tb]
\begin{center}
\begin{small}
\begin{tikzpicture}[node distance = 0.35cm, auto]
\node [inner sep=0pt] (a1) {\strut 中国 };
\node [inner sep=0pt, right of = a1, node distance = 4.5em] (ra1) {};
\node [inner sep=0pt, below of = a1, node distance=1.0em, font=\scriptsize] (ba1) {China};
\node [inner sep=0pt, below of = ba1, node distance=1.0em, font=\scriptsize] (ba2) {(1, 0)};
\node [inner sep=0pt, above of = a1, xshift=1.4em, yshift=0.7em,node distance=1.0em, font=\scriptsize,color=gray] (aa1) {SBJ};

\node [inner sep=0pt, above of = ra1, node distance =3.0em] (a2){\color{red} {\strut 鼓励} };
\node [inner sep=0pt, below of = a2, node distance=1.1em, font=\scriptsize] (ba2) {(0, 0)};
\node [inner sep=0pt, above of = a2, node distance=1.1em, font=\scriptsize] (bba2) {encourage};

\node [inner sep=0pt, right of = ra1, node distance = 2.5em] (a3) {\strut 外商};
\node [inner sep=0pt, below of = a3, node distance=1.0em, font=\scriptsize] (ba3) {foreign  merchant};
\node [inner sep=0pt, below of = ba3, node distance=1.0em, font=\scriptsize] (ba3) {(1, 0)};
\node [inner sep=0pt, above of = a3, xshift=-2.4em, yshift=0.1em,node distance=1.0em, font=\scriptsize,color=gray] (aa1) {COMP};

\node [inner sep=0pt, right of = a3, node distance = 5.5em, yshift=1.4em] (a4) {\strut 投资};
\node [inner sep=0pt, above of = a4, xshift=-2.5em, yshift=0.1em,node distance=1.0em, font=\scriptsize,color=gray] (aa1) {COMP};
\node [inner sep=0pt, below of = a4, node distance=1.0em, font=\scriptsize] (ba4) {invest};
\node [inner sep=0pt, below of = ba4, node distance=1.0em, font=\scriptsize] (bba4) {(1, 0)};

\node [inner sep=0pt, below of = bba4, node distance = 2.4em] (a5) {\color{red} {\strut 农业} };
\node [inner sep=0pt, below of = a5, node distance=1.0em, font=\scriptsize] (ba5) {agriculture};
\node [inner sep=0pt, below of = ba5, node distance=1.0em, font=\scriptsize] (bba5) {(2, 0)};
\node [inner sep=0pt, above of = a5, xshift=1.4em, yshift=0.4em,node distance=1.0em, font=\scriptsize,color=gray] (aa1) {COMP};


\path [draw, black, thick, ->] (a2) to (a1) {};
\path [draw, black, thick, ->] (a2) to (a3) {};
\path [draw, black, thick, ->] (a2) to (a4) {};
\path [draw, black, thick, ->] (bba4) to (a5){};
\end{tikzpicture}
\end{small}
\caption{The syntactic dependency tree of the sentence ``中国鼓励外商投资农业" (China encourages foreign merchants to invest in agriculture).
Numbers in brackets are the \textsc{DepPath} for each candidate argument with the predicate ``鼓励 (encourage)". Light grey labels on the arcs are the syntactic dependency labels.}
\label{graph:pathExample}
\end{center}
\end{figure}
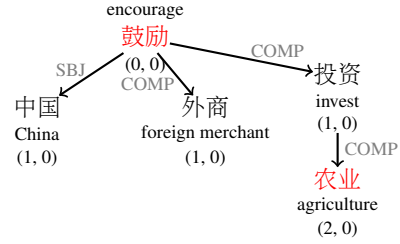

\begin{CJK}{UTF8}{gbsn}
As shown in Figure \ref{graph:pathExample}, the lowest common ancestor of the predicate ``鼓励 (encourage)" and the candidate argument ``农业 (agriculture)" is ``鼓励 (encourage)", so their \textsc{DepPath} is ``2,0" and its \textsc{RelPath} is ``COMP\_COMP,"\footnote{ When predicate is the ancestor of argument, sub-path from ancestor to predicate is none. We use `0' to represent distance and use empty string to represent the label path, and vice versa. }.
\end{CJK}

We take both \textsc{DepPath} and \textsc{RelPath} as common strings and map them into dense vectors in the similar way of \textsc{Dep} and \textsc{Rel}.

\begin{CJK}{UTF8}{gbsn}
\end{CJK}

\subsection{Incorporation Methods}

\subsubsection{Input Embedding Concatenation}
To incorporate syntactic knowledge, one simple method is to take it as part of the neural network input, denoted as \textsc{Input}.
We represent the syntactic information with dense vectors, and concatenate it with other information like word embedding: 
\begin{equation}
input =  E_W \oplus E_S. 
\end{equation}
where $\oplus$ means concatenation; $E_W$ means the original inputs of the neural model and $E_S$ means the embedding of syntax information, such as \textsc{Dep}/\textsc{Rel} or  \textsc{DepPath}/\textsc{RelPath}. 

\subsubsection{LISA}\label{lisa_description}

\label{lisaintro}
\citet{lisa2018} propose the linguistically-informed self-attention model (\textsc{LISA} for short) to combine SRL and dependency parsing as multi-task learning in a subtle way.
Based on the multi-head self-attention model, \textsc{LISA} uses one attention head to predict the dependency results and it can also directly use pre-trained dependency head results to replace the attention matrix  during testing.

\begin{figure}
  \centering
  \includegraphics[width=.48\textwidth]{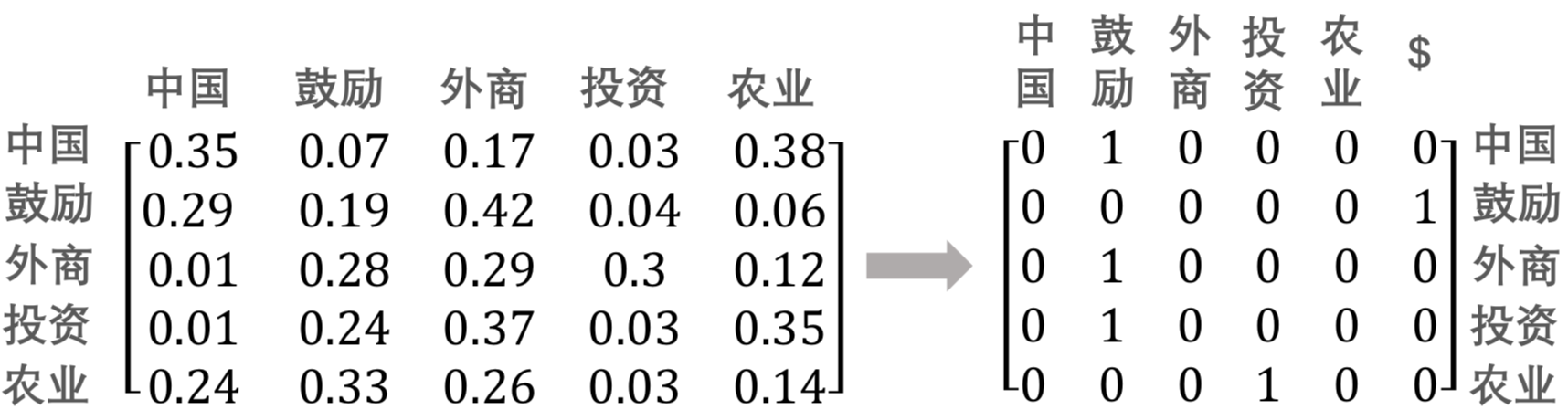} 
  \caption{Attention matrix of the replaced attention head in the \textsc{LISA} model. The left matrix is the original $softmax$ attention, and the right is a one-hot matrix copied from the syntactic dependency head results.  }
  \label{fig:lisa}
\end{figure}

Being different from their multi-task learning, we make the replacement of one attention head during both training and testing. Instead of the original $softmax$ attention matrix, we use a one-hot matrix, generated by mapping the dependency head index of each word into a 0-1 vector of the sentence length as Figure \ref{fig:lisa} shows.

We add the dependency relation information with $V$ in the replaced head so that we can make full use of the syntactic knowledge. 
The replaced attention head is formulated as follows:
\begin{equation}
Attention(Q, K, V) =  M_D \   (V \oplus E_R)
\end{equation}
where $M_D$ is the one-hot dependency head matrix and $E_R$ means the embedding of dependency relation information, such as \textsc{Rel} or \textsc{RelPath}. 

\subsubsection{Relation-Aware Self-Attention}
\label{relationaware}
Relation-aware self-attention model (\textsc{RelAwe} for brevity) incorporates external information into the attention. 
By this way, the model considers the pairwise relationships between input elements, which highly agrees with the task of SRL, i.e., aiming to find the semantic relations between the candidate argument and predicate in one sentence. 

Compared to the standard attention, in this paper, we add the dependency information into $Q$ and $V$ in each attention head, like equation (\ref{ra}) shows: 
\begin{equation}
\small
\begin{aligned}
\label{ra}
\scriptsize
A&ttention(Q, K, V)  =  \\
     &softmax(\frac{(Q + E_D + E_R) K^T}{\sqrt{d_k}}) (V+E_D+E_R)
\end{aligned}
\end{equation}
where $E_D$ and $E_R$ mean the syntactic dependency head and relation information respectively.
For our multi-layer multi-head self-attention model, we make this change to each head of the first $N$ self-attention layers.

\section{Experiment}

\begin{table}[t]
\small
\begin{center}
\begin{tabular}{l | r   r | r   r}
\hline   &  \multicolumn{2}{c|}{\textbf{dev}}  &\multicolumn{2}{c}{\textbf{test}}  \\ 
 &  \textbf{UAS} & \textbf{LAS} & \textbf{UAS} & \textbf{LAS}  \\ \hline
\hline \textsc{Auto} &   80.50  &  78.34  &    80.70  &  78.46      \\
\hline \textsc{Biaffine} &    89.00  &  85.86  &  89.05  &  85.60     \\
\hline \textsc{BiaffineBert} &    91.76  &  89.08  &  92.14  &  89.23    \\ \hline
\end{tabular}
\end{center}
\caption{Syntactic dependency performance for different parsers. \textsc{Auto} indicates the automatic dependency trees provided by the CoNLL-09 Chinese dataset. \textsc{Biaffine} means the trees are generated by BiaffineParser with pre-trained word embedding on the Gigaword corpus while \textsc{BiaffineBert} is the same parser with BERT. We use the labeled accuracy score (LAS) and unlabeled accuracy score (UAS) to measure the quality of syntactic dependency trees.}
\label{dep_res}
\end{table}

\subsection{Settings}

\textbf{Datasets \& Evaluation Metrics}
Our experiments are conducted on the CoNLL-2009 shared task dataset \cite{hajic2009}. 
We use the official evaluation script to compare the output of different system configurations, and report the labeled precision (P), labeled recall (R) and labeled f-score (F1) for the semantic dependencies.

\textbf{Word Representations}
Most of our experiments are conducted in the \emph{closed} setting without any external word embeddings or data resources than those provided by the CoNLL-2009 datasets. 
In the \emph{closed} setting, word embedding is initialized by a Gaussian distribution with mean $0$ and variance $\frac{1}{\sqrt{d}}$, where $d$ is the dimension of embedding size of each layer.

For the experiments with external resources in the \emph{open} setting, we utilize 1) word embeddings pre-trained with GloVe \cite{pennington2014glove} on the Gigaword corpus for Chinese and the published embeddings with 100 dimensions pre-trained on Wikipedia and Gigaword for English; and 2)
ELMo\footnote{We use the released model on their website: https://github.com/allenai/allennlp/blob/master/tutorials /how\_to/elmo.md} \cite{peters2018deep} and BERT\footnote{We generate our pre-trained  BERT embedding with the released model in https://github.com/google-research/bert. The model uses character-based tokenization for Chinese, which  require us to maintain alignment between our input text and output text of Bert. So we take take embedding of the first word piece as the whole word representation.} \cite{devlin2018bert}, two recently proposed effective deep contextualized word representations\footnote{In \emph{open} setting, we use pre-trained word embedding instead of random initialized embedding. If using BERT and Elmo additionally, we project them into representation vectors of same dimension as word embedding and concatenate them with other input.}.

Other embeddings, i.e., POS embedding, linguistic knowledge embedding, and so on are initialized in same way as random word embedding no matter in \emph{closed} or \emph{open} setting.

\textbf{Syntactic Parsers}
In Table \ref{dep_res}, both \textsc{Auto} and \textsc{Gold} syntactic dependencies are provided by the dataset.
Since the performance of the \textsc{Auto} is far behind the state-of-the-art BiaffineParser \cite{dozat2016deep}, we generate more dependency results by training BiaffineParser\footnote{We split train data into 5-fold, and train model with 4-fold to generate automatic trees of the left 1-fold train data.} with different external knowledge, including pre-trained word embedding and BERT.
Performance for different parsers is listed in Table \ref{dep_res}.

\textbf{Parameters} In this work, we set word embedding size $d_w=100$, POS embedding size $d_t=50$. The predicate embedding size is set as $d_p=100$. The syntax-related embedding size varies along with different configurations, so as the feature embedding size $d_f$. 

To facilitate residual connections, all sub-layers in the model produce outputs of dimension $d_{model}=d_f+d_p$. 
The hidden dimension $d_{ff}=800$ is applied for all the experiments. 
We set the number of shared self-attention blocks $N=10$. 
The number of heads varies with $d_{model}$, but dimension of each head is 25. 
Besides, \textsc{LISA} incorporates syntax knowledge in the 5-th self-attention layer while \textsc{RelAwe} incorporates in the first 5 layers.

We apply the similar dropout strategy as \citet{vaswani2017}, i.e., the  attention and residual dropout values are $0.2$ and $0.3$ respectively. 
The dropout is also applied in the middle layer of FFN with value $0.2$.
We also employ label smoothing \cite{Szegedy-layernorm} of value $0.1$ during training.

We use softmax-cross-entropy as our loss function, and use the Adadelta optimizer \cite{Zeiler-adadelta} with $\epsilon=10^{-6}$ and $\rho=0.95$. For all experiments, we train the model $200,000$ steps with learning rate $lr=1.0$, and each  batch has $4096$ words. 

All the hyper-parameters are tuned on the development set.

\begin{table}[t]
\small
\begin{center}
\begin{tabular}{l | l}
\hline  \textbf{Abbreviation}  & \textbf{Description}  \\ \hline
\ \ \textsc{None} & No syntactic knowledge \\ \hline 
\textbf{Syn. Parser} & \\
\ \ \textsc{Auto} &  Parsing result from CoNLL-2009  \\
\ \ \textsc{AutoDel} & Remove syntax errors of \textsc{Auto}   \\
\ \ \textsc{Biaffine} & BiaffineParser result  \\
\ \ \textsc{BiaffineBert} & BiaffineParser with BERT  \\
\ \ \textsc{Gold} &  Gold syntax from CoNLL-2009  \\ \hline
\textbf{Syn. Representation} & \\
\ \ \textsc{Dep} &  Dependency head \\
\ \ \textsc{Rel} & Dependency relation label  \\
\ \ \textsc{DepPath} & Tree-based position feature  \\
\ \ \textsc{RelPath} & Shortest relation label path  \\ \hline
\textbf{Incorporation} & \\
\ \ \textsc{Input} &  Add to the input embedding  \\ 
\ \ \textsc{LISA} & From \citet{lisa2018} \\
\ \ \textsc{RelAwe} & Relation-aware self-attention  \\ \hline
\end{tabular}
\end{center}
\caption{A glossary of abbreviations for different system configurations in our experiments.}
\label{glossary_table}
\end{table}

\textbf{Configurations}
We use different abbreviations to represent the parsing results, syntactic dependency representations, and incorporation methods. All the system configurations in our experiments are listed in Table \ref{glossary_table}.

\begin{table}[t]
\small
\begin{center}
\begin{tabular}{l | c | c | c}
\hline   & \textbf{P}  & \textbf{R} & \textbf{F1}  \\ \hline
\textsc{None} & 83.97 & 82.94 & 83.45 \\ \hline 
\textsc{Auto} & 85.92 & 84.62 & 85.26 \\ 
\textsc{Biaffine} & 86.04 & 85.43 & 85.73 \\ 
\textsc{BiaffineBert} & 87.32 & 87.10 & \textbf{87.21} \\ \hline 
\textsc{AutoDel} & 88.33 & 87.67 & 88.00 \\  
\textsc{Gold} & 91.00 & 91.43 & 91.22 \\ \hline 
\end{tabular}
\end{center}
\caption{SRL results with dependency trees of different quality on the Chinese dev set. These experiments are conducted on the \textsc{RelAwe} model with \textsc{Dep\&Rel} representations.}
\label{dep_srl_res}
\end{table}

\subsection{Quality of the Syntactic Dependencies}
We use the above-mentioned dependency trees of different quality for comparison, with \textsc{Dep\&Rel} representation on our \textsc{RelAwe} model.
In addition, we generate one more data \textsc{AutoDel} by deleting all the erroneous dependency heads and relations from the provided \textsc{Auto} data according to the gold heads and relations, and we do not replace them with any alternative heads and relations\footnote{For the AUTODEL data, we cannot guarantee that there exists a syntactic path between the two words. Therefore, we do not conduct experiments under the DepPath\&RelPath setting.}.
We take this setting as another reference (along with GOLD) to indicate that erroneous syntax information may hurt the performance of the SRL model.
We take the \textsc{Gold} as the upperbound reference of our task setting. Experiment results in Table \ref{dep_srl_res} demonstrate that, incorporating syntactic knowledge into the SRL model can achieve better performance and overall, the better the quality is, the better the SRL model performs. 
This is consistent with the previous study by \citet{he2017} on the English dataset.

Closer observation reveals two additional interesting phenomena. 
Firstly, SRL performance improvement is not proportionate to the improvement of dependency quality. 
When switching syntactic dependency trees from \textsc{Auto} to \textsc{Biaffine}, SRL performance improves 0.5\%, although syntactic dependency improves about 8\%. In contrast, the difference between \textsc{Biaffine} and \textsc{BiaffineBert} shows more significant improvement of 1.5\%.
The possible reason is that \textsc{BiaffineBert} provides key dependency information which is missing in other configurations. 
Secondly, the SRL performance gap between \textsc{AutoDel} and \textsc{Auto} is large though they provide the same correct syntactic information. 
This may indicate that incorporating erroneous syntactic knowledge hurts the SRL model, and even providing more correct dependencies cannot make up for the harm (cf. \textsc{BiaffineBert}).

\subsection{Representation of the Syntactic Dependencies}
\begin{table}[t]
\small
\begin{center}
\begin{tabular}{l | c | c | c}
\hline  & \textbf{P}  & \textbf{R} & \textbf{F1}  \\ \hline
\textbf{\textsc{Auto}}  &  &  &  \\
\ \ \textsc{DepPath\&RelPath} & 84.76 & 81.85 & 83.28 \\ \hline 
\textbf{\textsc{Biaffine} } &  &  &  \\
\ \   \textsc{Dep}  & 84.33 & 84.47 & 84.40 \\ 
\ \  \textsc{Rel} & 85.84 & 85.23 & 85.54 \\   
\ \  \textsc{Dep\&Rel} & 86.04 & 85.43 & 85.73 \\ 
\ \   \textsc{DepPath} & 85.48 & 84.17 & 84.82 \\  
\ \  \textsc{RelPath} & 86.85 & 83.92 & 85.36 \\ 
\ \  \textsc{DepPath\&RelPath}  & 86.40 & 85.52  & \textbf{85.96} \\ \hline 
\textbf{\textsc{Gold}}  &  &  &  \\
\ \ \textsc{DepPath\&RelPath} & 92.20 & 92.53 & 92.37 \\ \hline 
\end{tabular}
\end{center}
\caption{SRL results with different syntactic representations on the Chinese dev set. Experiments are conducted on the \textsc{RelAwe} method.}
\label{dp_dep_srl_res}
\end{table}

Apart from \textsc{Dep} and \textsc{Rel}, we also use \textsc{DepPath} and \textsc{RelPath} to encode the syntactic knowledge. In this subsection, we conduct experiments to compare different syntactic encoding in our SRL model. We base the experiments on our \textsc{RelAwe} model, since it is easier to incorporate different representations for comparison. When generating the \textsc{RelPath}, we filter the paths 1) when the dependency distance between the predicate and the candidate argument is more than 4, and 2) when the \textsc{RelPath}'s frequency is less than 10\footnote{As we know, dependency parsers cannot deal with long distance dependency well and it is unlikely to deliver a reliable result. And our experiments show that filtration achieves empirically better results.}.

No matter in which representation, dependency label information is more important than the head and the combination of the two achieves better performance as our experiment results in Table \ref{dp_dep_srl_res} show. 
Furthermore, using \textsc{Biaffine} dependency trees, \textsc{DepPath} and \textsc{RelPath} perform better than \textsc{Dep} and \textsc{Rel}.
This is because of the capability of \textsc{DepPath} and \textsc{RelPath} to capture more structural information of the dependency trees.

Comparing Table \ref{dep_srl_res} and \ref{dp_dep_srl_res}, when using gold dependencies, \textsc{DepPath\&RelPath} can achieve much better result than \textsc{Dep\&Rel}. 
But with the \textsc{Auto} trees, \textsc{DepPath\&RelPath} is much worse.  
Therefore, structural information is much more sensitive to the quality of dependency trees due to error propagation.

\subsection{Incorporation Methods}
 
\begin{table}[t]
\small
\begin{center}
\begin{tabular}{l | c | c | c}
\hline   & \textbf{P}  & \textbf{R} & \textbf{F1}  \\ \hline
\textbf{\textsc{Input} }  &  &  &  \\  
\ \ \textsc{Dep}  & 83.89 & 83.61 & 83.75 \\  
\ \ \textsc{Dep\&Rel} & 86.21 & 85.00 & 85.60 \\  
\ \ \textsc{Dep\&RelPath} & 86.01 & 85.38 & 85.69 \\ 
\ \  \textsc{DepPath\&RelPath} & 85.84 & 85.54 & \textbf{85.69} \\ \hline  

\textbf{\textsc{LISA} }  &  &  &  \\
\ \ \textsc{Dep} & 84.68 & 85.38 & 85.03 \\  
\ \ \textsc{Dep\&Rel} & 85.56 & 85.89 & 85.73 \\ 
\ \ \textsc{Dep\&RelPath}  & 85.84 & 85.64 & \textbf{85.74} \\  
\ \  \textsc{DepPath\&RelPath}\footnotemark[9]   & na & na & na \\ \hline

\textbf{\textsc{RelAwe} }  &  &  &  \\
\ \   \textsc{Dep}  & 84.33 & 84.47 & 84.40 \\ 
\ \   \textsc{Dep\&Rel} & 86.04 & 85.43 & 85.73 \\ 
\ \ \textsc{Dep\&RelPath}  & 86.21 & 85.01 & 85.60 \\ 
\ \  \textsc{DepPath\&RelPath}  & 86.40 & 85.52  & \textbf{85.96} \\ \hline 

\end{tabular}
\end{center}
\caption{
SRL results with different incorporation methods of the syntactic information on the Chinese dev set. Experiments are conducted on the \textsc{Biaffine} parsing results.}
\label{incoporate_way_res}
\end{table}
\footnotetext[9]{From the mechanism of \textsc{LISA}, we can find that the replaced attention head can't copy the syntactic dependency heads from \textsc{DepPath}.  }

\begin{table}[t]
\small
\begin{center}
\begin{tabular}{l | c | c | c}
\hline   &\textbf{P}  & \textbf{R} & \textbf{F1}  \\ \hline
\textbf{\textsc{Biaffine}}  &  &  &  \\
\ \ \ \ \textsc{Random}  & 86.40 & 85.52   & 85.96 \\   
\ \ \ \ \textsc{Giga} & 86.73 & 85.58 & 86.15 \\ 
\ \ \ \ \textsc{Elmo} & 86.84 & 86.74 & 86.79 \\  
\ \ \ \ \textsc{Bert} & 88.16 & 89.57 & \textbf{88.86} \\ \hline 
\textbf{\textsc{BiaffineBert}}  &  &  &  \\
\ \ \ \ \textsc{Bert}  & 88.05 & 89.65 & 88.84 \\ \hline 
\end{tabular}
\end{center}
\caption{SRL results with different external knowledge on the Chinese dev set. We use the \textsc{RelAwe} model and \textsc{DepPath\&RelPath} syntax representation.}
\label{external_know_res}
\end{table}

\begin{table*}[t]
\begin{center}
\begin{tabular}{l | c | c | c}
\hline  Chinese  & \textbf{P}  &  \textbf{R} & \textbf{F1}  \\ \hline
\textbf{\textsc{None}} & 81.99 & 80.65 & 81.31 \\ \hline
\textbf{Closed}  &  &  &  \\
\  \ \ CoNLL09 SRL Only & na & na & 78.6  \\ 
\  \ \ \textsc{Input(\textsc{DepPath\&RelPath})}  & 84.19 & 83.65 & 83.92 \\
\  \ \ \textsc{LISA(\textsc{Dep\&RelPath})}  & 83.84 & 83.54 & 83.69 \\
\  \ \  \textsc{RelAwe(\textsc{DepPath\&RelPath})}  & 84.77 & 83.68 & \textbf{84.22} \\ \hline
\textbf{Open}  &  &  &  \\
\ \ \ \citet{marcheggiani2017b} & na & na & 82.5 \\
\ \ \ \citet{cai2018} & 84.7 & 84.0  & 84.3 \\
\ \ \ \textsc{Input(\textsc{DepPath\&RelPath}) + \textsc{ BERT}}  & 86.89  & 87.75  & 87.32  \\ 
\ \ \ \textsc{LISA(\textsc{Dep\&RelPath}) + \textsc{ BERT}}  & 86.45  & 87.90 & 87.17  \\ 
\ \ \ \textsc{RelAwe(\textsc{DepPath\&RelPath}) + \textsc{ BERT} }  & 86.73 & 87.98 & \textbf{87.35}  \\ \hline
\textbf{\textsc{Gold}}  & 91.93 & 92.36 & 92.14 \\ \hline
\end{tabular}
\end{center}
\caption{SRL results on the Chinese test set. We choose the best settings for each configuration of our model.}
\label{sota_res}
\end{table*}


This subsection discusses the effectiveness of different incorporation methods of the syntactic knowledge. We take \textsc{Biaffine}'s output as our dependency information for the comparison.

Firstly, results in Table \ref{incoporate_way_res} show that with little dependency information (\textsc{Dep}), \textsc{LISA} performs better, while incorporating richer syntactic knowledge (\textsc{Dep\&Rel} or \textsc{Dep\&RelPath}), three methods achieve similar performance.
Overall, \textsc{RelAwe} achieves best results given enough syntactic knowledge.

Secondly, \textsc{Input} and \textsc{LISA} achieve much better performance when we combine the dependency head information and the relation, while \citet{lisa2018} have not introduced relation information to the \textsc{LISA} model and \citet{xia2018} have not combined the head and relation information either. Our proposed \textsc{RelAwe} method with \textsc{DepPath\&RelPath} representation performs the best, which encodes the richest syntactic knowledge.

Lastly, under the same settings, \textsc{LISA} and \textsc{RelAwe} perform better than \textsc{Input}, which indicates the importance of the location where the model incorporates the syntax, the input layer vs. the encoder layer.






\subsection{External Resources}

Apart from the experiments with syntactic knowledge itself, we also compare different external resources to discover their relationship with the syntax, including pre-trained word embeddings, ELMo, and BERT.
We conduct experiments with our best setting, the \textsc{RelAwe} model with \textsc{DepPath \& RelPath} and the results are listed in Table \ref{external_know_res}.

The plain word embedding improves a little in such settings with syntactic information, while for the newly proposed \textsc{Elmo} and \textsc{Bert}, both of them can boost the models further.

\subsection{Final Results on the Chinese Test Data}

Based on the above experiments and analyses, we present the overall results of our model in this subsection. We train the three models (\textsc{Input}, \textsc{LISA}, and \textsc{RelAwe}) with their best settings without any external knowledge as \textsc{Closed}, and we take the same models with \textsc{Bert} as \textsc{Open}. The \textsc{DepPath\&RelPath} from \textsc{Gold} without external knowledge serves as the \textsc{Gold} for reference.
Since we have been focusing on the task of argument identification and labeling, for both \textsc{Closed} and \textsc{Open}, we follow \citet{roth2016} to use existing systems' predicate senses \cite{johansson2008effect} to exclude them from comparison.

Table \ref{sota_res} shows that our \textsc{Open} model achieves more than 3 points of f1-score than the state-of-the-art result, and \textsc{RelAwe} with \textsc{DepPath\&RelPath} achieves the best in both \textsc{Closed} and \textsc{Open} settings.
Notice that our best \textsc{Closed} model can almost perform as well as the state-of-the-art model while the latter utilizes pre-trained word embeddings.
Besides, performance gap between three models under \textsc{Open} setting is very small.
It indicates that the representation ability of \textsc{BERT} is so powerful and may contains rich syntactic information.
At last, the \textsc{Gold} result is much higher than the other models, indicating that there is still large space for improvement for this task.


\begin{table}
\begin{center}
\begin{small}
\begin{tabular}{l | c | c | c}
\hline  \textbf{English} & \textbf{P}  &  \textbf{R} & \textbf{F1}  \\ \hline
\ \textsc{LISA}(\textsc{Dep})\footnotemark[10]  & 89.26 &85.46 & 87.32  \\ 
\ \textsc{LISA}(\textsc{Dep\&RelPath})  & 89.74  & 85.38 & 87.51 \\ 
\ \textsc{Input}(\textsc{DepPath}) & 89.33 & 85.60 & 87.42 \\
\ \textsc{Input}\scriptsize(\textsc{DepPath\&RelPath}) & 89.70 & 85.48 & 87.54 \\ 
\ \textsc{RelAwe}\scriptsize(\textsc{DepPath\&RelPath}) & 89.55 &85.92 & \textbf{87.70} \\ \hline
\end{tabular}
\end{small}
\end{center}
\caption{SRL results on the English test set. We use syntactic dependency results generated by  BiaffineParser (On test set, syntactic performance is: UAS = 94.35\%, and LAS = 92.54\%, which improves about 6\% compared to automatic trees in CoNLL-2009.). }
\label{english_res}
\end{table}

\subsection{Results on the English Data}

We also conduct several experiments on the English dataset to validate the effectiveness of our approaches on other languages than Chinese and the results are in Table \ref{english_res}.
Although both configurations are not exactly the same as their original papers, we tried our best to reproduce their methods on the CoNLL2009 dataset for our comparison.
Overall, the results are consistent with the Chinese experiments, while the improvement is not as large as the Chinese counterparts. The \textsc{RelAwe} model with \textsc{DepPath\&RelPath} still achieves the best performance. Applying our syntax-enhanced model to more languages will be an interesting research direction to work on in the future.
\footnotetext[10]{We reimplement \textsc{LISA}  in \citet{lisa2018} as \textsc{LISA}(\textsc{Dep}), and \citet{xia2018}'s best \textsc{DepPath} approach as \textsc{Input}(\textsc{DepPath}). Therefore, we can compare with their work as fairly as possible. Other settings are the best configurations for their corresponding methods.}

\section{Conclusion and Future Work}
This paper investigates how to incorporate syntactic dependency information into semantic role labeling in depth.
Firstly, we confirm that dependency trees of better quality are more helpful for the SRL task. 
Secondly, we present different ways to encode the trees and the experiments show that keeping more (correct) structural information during encoding improves the SRL performance. 
Thirdly, we compare three incorporation methods and discover that our proposed relation-aware self-attention-based model is the most effective one.

Although our experiments are primarily on the Chinese dataset, the approach is largely language independent. Apart from our tentative experiments on the English dataset, applying the approach to other languages will be an interesting research direction to work on in the future.

\bibliographystyle{acl_natbib}
\bibliography{emnlp-ijcnlp-2019}

\end{document}